\documentclass[letterpaper, 10 pt, conference]{ieeeconf}  

\IEEEoverridecommandlockouts                              

\usepackage[utf8]{inputenc} 
\usepackage[T1]{fontenc}    
\usepackage{hyperref}       
\usepackage{url}            
\usepackage{booktabs}       
\usepackage{amsfonts}       
\usepackage{nicefrac}       
\usepackage{microtype}      
\usepackage{xcolor}         
\usepackage{graphicx}
\usepackage{multirow}
\usepackage{adjustbox}
\usepackage{caption}
\usepackage{amsmath}
\usepackage{lscape}
\usepackage{tablefootnote}

\newcommand{\equref}[1]{Eq ~\ref{#1}}
\newcommand{\figref}[1]{Figure ~\ref{#1}}

\title{\LARGE \bf
VL-Grasp: a 6-Dof Interactive Grasp Policy for Language-Oriented Objects in Cluttered Indoor Scenes
}

\author{%
    Yuhao Lu, Yixuan Fan, Beixing Deng, Fangfu Liu, Yali Li, Shengjin Wang\\
    Tsinghua University
}

\begin{document}

\maketitle
\thispagestyle{empty}
\pagestyle{empty}

\begin{abstract}
  Robotic grasping faces new challenges in human-robot-interaction scenarios. We consider the task that the robot grasps a target object designated by human's language directives. The robot not only needs to locate a target based on vision-and-language information, but also needs to predict the reasonable grasp pose candidate at various views and postures. 
  In this work, we propose a novel interactive grasp policy, named Visual-Lingual-Grasp (VL-Grasp), to grasp the target specified by human language.
  First, we build a new challenging visual grounding dataset to provide functional training data for robotic interactive perception in indoor environments. Second, we propose a 6-Dof interactive grasp policy combined with visual grounding and 6-Dof grasp pose detection to extend the universality of interactive grasping. Third, we design a grasp pose filter module to enhance the performance of the policy. Experiments demonstrate the effectiveness and extendibility of the VL-Grasp in real world. The VL-Grasp achieves a success rate of 72.5\% in different indoor scenes. The code and dataset is available at \url{https://github.com/luyh20/VL-Grasp}.

\end{abstract}


\section{Introduction}

With the development of deep learning methods, data-driven robotic grasping approaches have been greatly developed.
To further dig deeper into cognition and interaction ability of the robot, the researches of interactive robot manipulations are gradually emerging \cite{dobrev2016multi, liu2020self, noda2014multimodal, shridhar2020ingress, xompero2022audio, stepputtis2020language}.
In this work, we focus on the interactive grasp task of grasping a unique target specified by a human lingual command. 

The interactive grasp task has been focused on and continually studied. This task can be basically divided into two steps: finding the target by a natural language directive and predicting how to grasp it. 
Previous researches \cite{chen2021joint, ding2022visual, shridhar2018interactive, shridhar2020ingress, zhang2021invigorate} introduce the visual grounding task \cite{qiao2020referring} to achieve perception and positioning for the target in the first step, and predict the grasp pose with the help of 2D-plane grasp pose detection method in the second step. 

\begin{figure}[t]
  \centering
  \includegraphics[width=\linewidth]{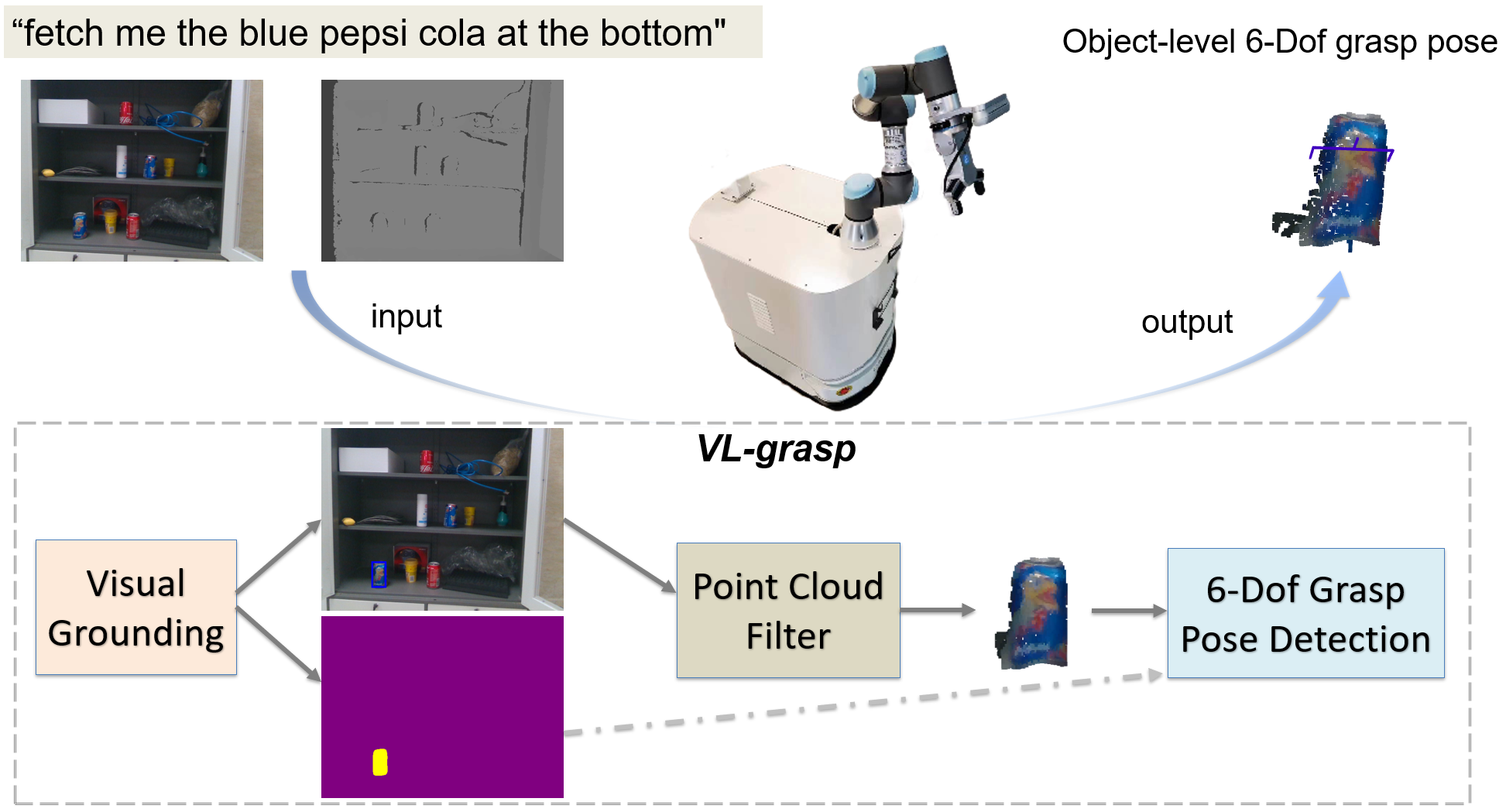}
  \caption{Overview. The VL-Grasp inputs an RGB image, a depth image and a natural language text, and outputs a optimal 6-Dof grasp pose. The VL-Grasp mainly contains three modules: a visual grounding network, a point cloud filter module, and a 6-Dof grasp pose detection network.}
  \label{fig:first}
\end{figure}

However, existing interactive grasp applications in real physical environment are limited by two problems. 
On the one hand, robotic observation angle and grasping scene are restricted in previous approaches \cite{chen2021joint, ding2022visual, shridhar2018interactive, shridhar2020ingress, zhang2021invigorate}. Because 2D-plane grasping place restrictions on gripper direction, the robot always looks from the top of the table and the object is always on the table. This restricts the practical application scope of robots. For example in \figref{fig:first}, previous approaches are not suitable to handle the grasping problem under a shelf scene. Therefore, 2D-plane grasping is hard to meet the demand of actual interactive grasp with a more diverse and complex environment.
On the other hand, the interactive grasp task lacks dedicated training data of the visual grounding. Previous studies \cite{shridhar2018interactive, shridhar2020ingress, zhang2021invigorate, chen2021joint} rely on the popular visual grounding datasets like RefCOCO \cite{yu2016modeling}, which has low correlation between data contents and robotic grasping. Popular datasets \cite{yu2016modeling, kazemzadeh2014referitgame, plummer2015flickr30k} usually contain few objects suitable for grasping, and few scenes from multiple robotic observation angles. Although some datasets \cite{chen2020scanrefer, wang2021ocid} collect indoor scenes, they lack ambiguous samples that there are multiple objects with the same category in an image. The training with ambiguous samples determines the ability of the model to distinguish ambiguous objects.

To tackle above problems, in this paper, we propose a novel interactive grasp policy to achieve multi-view and multi-scene interactive grasp, and establish a new challenging visual grounding dataset to provide functional data for it. 
First, we design a interactive grasp policy based on the visual grounding and the 6-Dof grasp pose detection method, called VL-Grasp. The overview of the VL-Grasp is shown in \figref{fig:first}. To extend the grasping space from single direction to any angle, the VL-Grasp innovatively applies the 6-Dof grasp pose detection approach into the interactive grasp task. With the help of 6-Dof grasp pose detection method, the interactive grasp can also adapt to various observation views and more diverse indoor scenes. To integrate grasping and perception process in 3D space, the VL-Grasp leverages a point cloud filter module to convert the scene-level depth image into the object-level point cloud. Real robot experiments are conducted to demonstrate the effectiveness and adaptability of our proposed policy. 
Second, we build a new visual grounding dataset for robot reasoning, called RoboRefIt. The RoboRefIt selects 66 categories graspable desktop objects that include some from robotic grasping datasets \cite{fang2020graspnet}. We set up a variety of indoor scenes and grasping space to place objects. Multiple experiments validate the functionality and usability of the dataset.

\begin{figure*}[t]
  \centering
  \includegraphics[width=\textwidth]{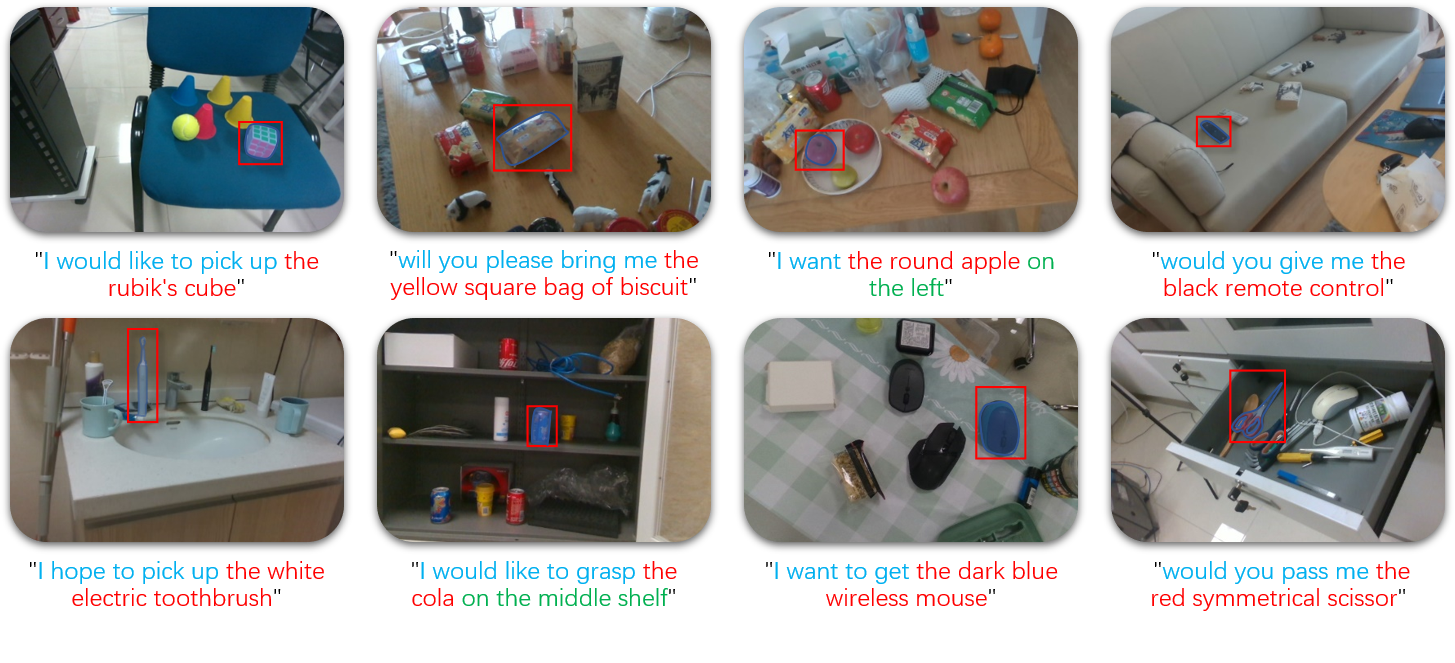}
  \caption{Scenario examples of the RoboRefIt. The red boxes in the pictures show the corresponding target of referring expressions. In the expressions given by the \equref{eq1}, the blue part belongs to the beginning template; the red part belongs to the attribution and object; the green part belongs to the location.}
  \label{fig:demo}
\end{figure*}

In summary, the main contributions of this paper are as follows: 
\begin{itemize}
\item We propose a novel interactive grasp policy, VL-Grasp. The VL-Grasp innovatively combines visual grounding and 6-Dof grasp pose detection method to handle the interactive grasp task. This policy expands the practical application space and application value of interactive grasp.

\item We contribute a new challenging visual grounding dataset, RoboRefIt, used for robot visual-lingual reasoning. 

\item We designs a point cloud filter module that improves grasping success rate in complex scenes. 
\end{itemize}

\section{Related Works}
\label{gen_inst}

\paragraph{\textbf{Visual Grounding}}
Visual grounding addresses the problem of locating a target object in an image given a sentence. 
Generally, the location result with bounding box is called referring expression comprehension (REC), and the location result with object mask is called referring expression segmentation (RES). The pipeline of visual grounding methods extract visual features and language features individually, and aggregate the cross-modal features to predict the bounding box or the segmentation mask of the target. Traditional methods \cite{yu2018mattnet, nagaraja2016modeling, hu2016natural, zhang2018grounding, shi2018key, jing2021locate} use the convolutional neural networks \cite{simonyan2014very} and long short-term memory \cite{greff2016lstm} (CNN-LSTM) framework to learn features. Recently, the transformer mechanism \cite{vaswani2017attention} is widely used in the REC and RES methods \cite{ye2019cross, deng2021transvg, liu2021cross, yang2022improving, feng2021encoder}, which is considered to be conducive to cross-modal learning. In the field of visual grounding, the REC and RES tasks are usually studied separately. However, since these tasks are consistent in the need to capture the long-range dependencies between linguistic and visual features, some unified framework \cite{luo2020multi, li2021referring} are also proposed by leveraging shared backbone and task-specific heads.

\paragraph{\textbf{Visual Grounding Datasets}}
Many datasets \cite{qiao2020referring} have been constructed for researches of visual grounding in the computer vision community. Some early datasets \cite{mitchell2013generating, kazemzadeh2014referitgame, de2017guesswhat, mao2016generation, plummer2015flickr30k} mainly generate referring expressions by manual annotation. The ReferItGame \cite{kazemzadeh2014referitgame} produces expressions by a two-player game. The RefCOCO and RefCOCO+ are based on the MSCOCO \cite{lin2014microsoft} and collect language by using the annotation rule \cite{kazemzadeh2014referitgame} as well. 
Recent datasets \cite{hudson2019gqa, karpathy2015deep, mauceri2019sun, chen2020scanrefer} adopt automatic annotation methods, which generate expression indirectly with the help of scene graph annotation or structured template synthesis.
The Cops-ref \cite{chen2020cops} is built on top of the real-world images in GQA\cite{hudson2019gqa} and generates referring expressions by utilising the scene graph annotations and further data cleaning. The ScanRefer \cite{chen2020scanrefer} and the SUNRefer \cite{liu2021refer} annotate 3D bounding boxes for the targets. The OCID-Ref \cite{wang2021ocid} is a visual grounding dataset for robotics utilizing the images and object labels from OCID \cite{richtsfeld2012segmentation}.

\paragraph{\textbf{Grasp Pose Detection} }
Grasp pose detection (GPD) can be divided into 2D-plane grasping and 6-Dof grasping \cite{yin2022overview}. The 2D-plane grasping limits the posture of the gripper perpendicular to the desktop. The 6-Dof GPD methods has been studied more extensively in recent years. Many recent methods \cite{liang2019pointnetgpd, murali20206, sundermeyer2021contact, ten2017grasp, Wang_2021_ICCV, fang2022anygrasp, chao2021dexycb} are mainly based on deep learning. GraspNet-1Billion \cite{fang2020graspnet} builds a large-scale grasp dataset and proposes a baseline method for learning grasp poses. Zhao et al. \cite{zhao2020regnet} use group region features to predict grasp proposals. Qian et al.\cite{qian2020grasp} propose a new single-view approach by linking affordances-based task constraints to the grasp pose. Dex-nerf \cite{ichnowski2021dex} combines a neural radiance field \cite{mildenhall2021nerf} and DexNet \cite{mahler2017dex} to grasp transparent objects. Lou et al. \cite{lou2021collision} propose a novel deep neural network to grasp novel objects by self-supervision training in simulation.

\section{RoboRefIt Dataset}
Existing visual grounding datasets are hard to be served as a suitable practical test bed for human-robot
interaction because of their low consistency with robotics. In this paper, we establish a new challenging visual grounding dataset for robotic perception and reasoning in indoor environments, called RoboRefIt. 
The RoboRefIt aims to help robots find the target according to a language command. The RoboRefIt comprises 10,872 RGB-D images and 50,758 referring expressions. All of these images are collected in real-world indoor scenes, involving 66 object categories and 187 different scenes. And there are 5,636 images appearing multiple objects of the same kind. The RoboRefIt annotates two types of labels: 2D bounding box and segmentation mask. We disclose the RoboRefIt dataset at the homepage\protect\footnotemark[1]. In addition, statistics information and more details of the dataset can also be found on the homepage\protect\footnotemark[1].

\footnotetext[1]{\url{https://luyh20.github.io/RoboRefIt.github.io/}}

\subsection{Data collection}

\paragraph{Image collection.} We select 23 categories of objects in the GraspNet-1Billion dataset \cite{fang2020graspnet} and 43 categories of objects of our own as the object set. The categories of these objects include \emph{fruits, food, tools, toiletries, toy models} and so on. And these objects are easy for robot to grasp. Moreover, there are roughly six kinds of life scenes to place objects, including \emph{tables, chairs, shelves, wash tables, drawers and sofas}. The images are collected from 187 different cluttered scenes of daily life. Each scene selects 3 to 10 objects from object set. To increase the complexity of ambiguous samples, we set up two or three objects of the same category in multiple scenes. We place different objects and adjust the pose and position of objects in each scene, and make the camera collect pictures from different directions. Two Intel Realsense D435i cameras and one Intel Realsense D455 camera, are used to capture raw RGB and depth data. The RGB image and depth image will be aligned with the resolution of \emph{640$\times$480} during acquisition. Finally, each scene contains about 20 to 100 pictures.

\paragraph{Referring Expression Generation.} The purpose of referring expression generation (REG) is to create a unique semantic description for an object in a scene image. The visual grounding datasets actually build a large number of image-text pairs. And the REG is the key to the construction of the datasets. In the RoboRefIt, we utilize a REG engine to generate diverse, complex and unambiguous expressions to locate the corresponding objects. The REG engine is a semi-automatic text generation technology, combined with text synthesis and some manual annotation. We describe the robot directive into a basic form like \eqref{eq1}, where a complete sentence consists of four language modules as shown in \figref{fig:demo}.

\begin{equation}
    \begin{split}
    robot\ directive = <beginning\ template> + \\ 
    <attribution> + <object> + <location>
    \end{split}
    \label{eq1}
\end{equation}

We design 66 kinds of beginning templates of robot directives and manually annotate diverse and rich phrase embedding of attribution and object name for the object set. And we provide the location tags for objects with contextual ambiguity when multiple similar or identical objects appear in the scene. Then, each scene will be equipped with the corresponding phrase embedding of each module. The REG engine will randomly sample and combine the phrase embedding to output the final text descriptions. 

\subsection{Data annotation}
The RoboRefIt contains learning labels for two subdivided tasks: 2D bounding boxes for REC and segmentation masks for RES. Since the RGB images and depth images are aligned, all these annotation results are also applicable to the depth images. We employ experienced annotation experts from professional artificial intelligence data service provider to complete the labeling tasks. The collected images and text data are assigned to annotators. Annotators mark bounding boxes and masks of the objects in turn according to multiple text descriptions of each image. Meanwhile, due to some possible deficiencies in the process of referring expression generation, annotators need to supplement the location or attribute characteristics of the object until the expressions satisfy the unique unambiguous condition. The preliminary labeling results will go through at least 3 rounds of inspection by professional checkers. 

\begin{figure*}[t]
  \centering
  \includegraphics[width=\textwidth]{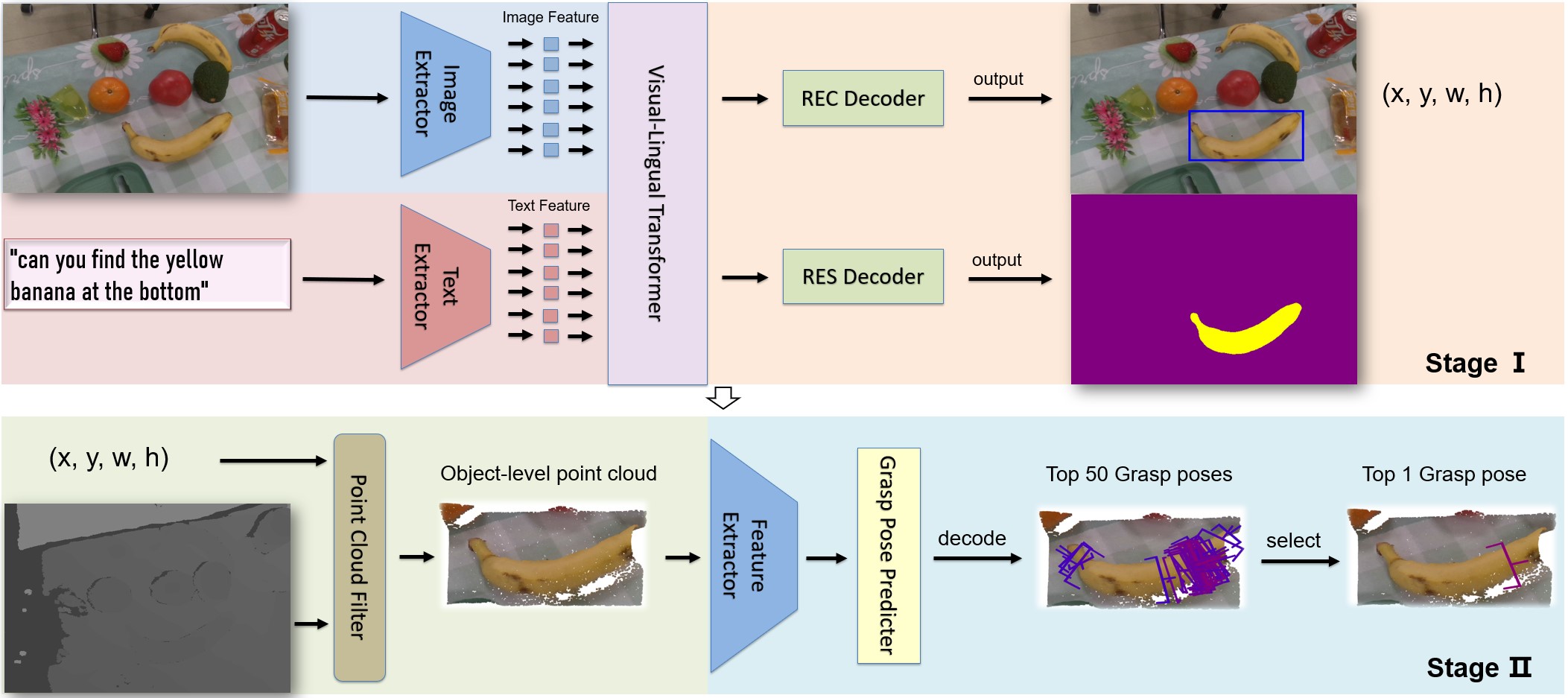}
  \caption{The overall architecture of VL-Grasp. The stage \uppercase\expandafter{\romannumeral1} uses a visual grounding network to locate the target with RGB image and text inputs. The stage \uppercase\expandafter{\romannumeral2} combines a point cloud filter and a 6-Dof grasp pose detection network to predict optimal grasp pose.}
  \label{fig:overview}
\end{figure*}

\subsection{Comparison with Existing Datasets}
The RoboRefIt differs existing visual grounding datasets mainly in the following aspects: 1) Content relevance with robot (indoor scenes, graspable objects);
2) Format of annotation and image data;
3) Scale of expressions and images;
4) Diversity of objects; 
5) Ambiguous samples.
Key characteristics are summarized in Table \ref{tab:my-table}. 
Popular datasets \cite{yu2016modeling, mao2016generation, kazemzadeh2014referitgame, plummer2015flickr30k} usually include few human-robot interaction indoor scenes and contain only RGB image data.
Compared with the datasets \cite{chen2020scanrefer, mauceri2019sun, liu2021refer, wang2021ocid} that have indoor robot scenes, the RoboRefIt pays more attention to graspable desktop objects and cluttered scenes and is the largest dataset in terms of number of images. The indoor datasets \cite{chen2020scanrefer, mauceri2019sun, liu2021refer} don't annotate segmentation masks, which is very important for post-filter of 6-Dof grasp poses. Moreover, the RoboRefIt has a large mount of ambiguous samples by manual setting. But the indoor datasets \cite{chen2020scanrefer, mauceri2019sun, liu2021refer, wang2021ocid} are derived datasets, and their source datasets don't contain a lot of ambiguous samples.

\begin{table*}[]
\centering
\caption{Statistic Comparison of Previous Datasets and the RoboRefIt in Main Characteristics.}
\label{tab:my-table}
\resizebox{\textwidth}{!}{%
\begin{tabular}{cccccccccc}
\toprule
Dataset  & \begin{tabular}[c]{@{}c@{}}Indoor \\ Scene\end{tabular}  &\begin{tabular}[c]{@{}c@{}}Annotation\\ format\end{tabular}   &\begin{tabular}[c]{@{}c@{}}Data\\ format\end{tabular}  
& \begin{tabular}[c]{@{}c@{}}Number of \\ images\end{tabular} 
& \begin{tabular}[c]{@{}c@{}}Number of \\ expressions\end{tabular} 
& \begin{tabular}[c]{@{}c@{}}Number of \\ objects\end{tabular} 
& \begin{tabular}[c]{@{}c@{}}Object \\ categories\end{tabular} 
& \begin{tabular}[c]{@{}c@{}}Avg. \\ length words\end{tabular} \\ \hline
ReferItGame\cite{kazemzadeh2014referitgame}  &No &2D bbox  & RGB  & 19,894                                                                     & 130,525                                                          & 95,654                  &-                      & 3.61                                                         \\
RefCOCO\cite{yu2016modeling} &No    &2D bbox/mask  & RGB & 19,994                                                                        & 142,209                                                          & 50,000              &80                                         & 3.61                                                         \\
RefCOCO+\cite{yu2016modeling} &No  &2D bbox/mask &RGB    & 19,992                                                               & 141,564                                                          & 49,856                     &80                                  & 3.53                                                         \\
RefCOCOg\cite{mao2016generation} &No  &2D bbox/mask   & RGB & 26,771                                                                        & 104,560                                                          & 54,822                   &80                                    & 8.43                                                         \\
Flickr 30k entities\cite{plummer2015flickr30k} &No    & 2D bbox   &RGB     & 31,783                  & 158,915       &275,775   &44,518     & - \\

Cops-ref\cite{chen2020cops}  &No   & 2D bbox    &RGB     & 703                                                                  & 148,712                                                          & 1,307,885                         &508                           & 14.4                                                         \\

Sun-Spot\cite{mauceri2019sun} &Yes   &2D bbox   &RGBD     & 1,948                                                                        & 7,990                                                            & 3,245                           &38                               & 14.04                                                        \\
ScanRefer\cite{chen2020scanrefer} &Yes   &3D bbox  &3D       & 800         & 51,583                                                           & 11,046             &250                                             & 20.07                                                  \\
SUNRefer\cite{liu2021refer} &Yes   &3D bbox   &3D        & 7,699                                                                                & 38,495                                                           & 7,699                      &-                                      & 16.3                                                        \\
OCID-Ref\cite{wang2021ocid}  &Yes   &mask   &3D         & 2,300
& 267,339                                                & -                  &58                                          & 8.56                                                         \\ \hline
RoboRefIt(ours) &Yes    &2D bbox/mask  &RGBD   & 10,872                                                                                    & 50,758          
                & 50,758                     &66        & 9.52     \\
 \bottomrule
                
\end{tabular}%
}
\end{table*}

\section{VL-Grasp}
We propose the two-stage learning-based policy for the interactive grasp task, called VL-Grasp.

\subsection{Problem Statement}
Given an RGB image \emph{I}, a depth image \emph{D} a natural language command \emph{C} that refers to the only one object in the scene, the interactive grasp task aims to predict the optimal 6-Dof grasp pose configuration \emph{G} for the target object \emph{O}. This problem is mainly decoupled into two stages in the VL-Grasp. The first stage inputs the RGB image \emph{I} and the command \emph{C}, and outputs the position results of the target object that comprise the 2D bounding box \emph{B} and the segmentation mask \emph{M}. The second stage inputs the results and the depth image \emph{D}, and outputs the final grasp configuration \emph{G}.

\subsection{Overview}
The overview architecture of the VL-Grasp is depicted in \figref{fig:overview}. The VL-Grasp designs a two-stage pipeline and the network at each stage is trained separately. \textbf{Stage \uppercase\expandafter{\romannumeral1}} deals with the problems of perception and positioning. The visual grounding network inputs an RGB image and a natural language text, and outputs a 4D parameters $(x, y, w, h)$ of the bounding box and a mask image. The model is trained in our visual grounding dataset, RoboRefIt. \textbf{Stage \uppercase\expandafter{\romannumeral2}} is to further complete the reasoning and decision-making of grasping. First, the point cloud module inputs the 4D parameters and a depth image to obtain object-region point cloud. Second, the 6-Dof grasp pose detection inputs the object-region point cloud and outputs a 6D parameters $(x, y, z, rx, ry, rz)$ of optimal candidate grasp. The detailed contents of each stage are described in the following.

\subsection{Visual Grounding Network}
To obtain both bounding box and mask results of the object, the VL-Grasp adopts the unified visual grounding framework, Referring Transformer \cite{li2021referring}, in \textbf{Stage \uppercase\expandafter{\romannumeral1}}. Firstly, the language model BERT \cite{devlin2018bert} is used as the text extractor to extract text features of the natural language command \emph{C}. And the ResNet \cite{he2016deep} is used as the image extractor to extract image features of the RGB image \emph{I}. Secondly, a visual-lingual transformer structure fuses the features of vision and language features through the cross-modal attention mechanism. Thirdly, the visual-lingual features forward the REC and RES decoders to predict the bounding box \emph{B} and segmentation mask \emph{M} individually.

\subsection{Point Cloud Filter}
In \textbf{Stage \uppercase\expandafter{\romannumeral2}}, the VL-Grasp transform the scene-level point cloud to object-level point cloud via a point cloud filter at first. The point cloud filter crops the depth image with the dilated bounding box. Then, the cropped depth image is converted to single-view point cloud with camera internal parameters. The dilation operation expands the scale of the bounding box with a size parameter, and retains more complete object geometry information and edge background geometry information. In the experiments, we prove that the point cloud filter is beneficial to the interference of the grasp pose detection model and improves the performance of real robot grasping.

\subsection{6-Dof Grasp Pose Detection Network}
The VL-Grasp applies the multi-resolution learning-based network based on the FGC-GraspNet \cite{lu2022hybrid}, to detect multiple available 6-Dof grasp pose proposals in \textbf{Stage \uppercase\expandafter{\romannumeral2}}. The point cloud filter outputs the object-level point cloud, then the PointNet++ \cite{qi2017pointnet++} backbone is used as the feature extractor to extract point cloud features.  The grasp pose predictor inputs the point cloud features and outputs various characteristic parameters of grasp poses. The grasp pose predictor is mainly composed of MLP layers and local attention module, which is used to decode the point cloud features into characteristic parameters of grasp poses. Specifically, low-resolution features forward the MLP layers to predict the approach directions of candidates of grasp poses and foreground mask of the object. And High-resolution features forward the local attention module and MLP layers to predict the rotation direction and depth of candidates of grasp poses, and confidence scores of candidates. Then, these characteristic parameters of high-score candidates are decoded into the 6-Dof configurations of grasp poses. Finally, the VL-Grasp filters the grasp poses through the object mask and selects the highest-confidence candidate.

\section{Experiments}
In this section, we demonstrate the effectiveness and rationality of the RoboRefIt and show the performance of the VL-Grasp in real world. 

\subsection{Data Split}
\label{sub:split}
The RoboRefIt divides the original data into the train set, the testA set and the testB set.
During the actual deployment of the visual grounding tasks, robots are faced with different scenes of life. To evaluate adaptability of models with different scene distribution, the RoboRefIt sets up the testA set and the testB set that are different in the correlation of scene distribution with the training set.
The train set and testA set are split from one part selected scenes by stratified random sampling, which ensures the similarity of scene distribution between these two subsets. The testB set directly adopts another part selected scenes, which leads to the difference of scene distribution between the testB set and train set. For example, the testB set contains some different scenes that don't appear in the train set, such as a drawer or different tables.
Consequently, the testA set can be used to measure the basic prediction ability of the models and the testB set can be used to measure the generalization ability.

\subsection{REC and RES Experiments}

\begin{table}[t]
  \begin{minipage}{0.5\textwidth}
  \caption{REC and RES Results under the RefTR Model \\ on the RoboRefIt Dataset.}
  \label{two:experiment}
  \centering
  \begin{tabular}{c|c|c|cc}
    \toprule
    
    \multicolumn{1}{c}{\multirow{2}{*}{Task}} &
    \multicolumn{1}{|c|}{\multirow{2}{*}{Backbone}}&  
    \multicolumn{1}{|c|}{\multirow{2}{*}{Input}}&
    \multicolumn{2}{|c}{RoboRefIt} \\

    \cline{4-5}
    \multicolumn{1}{c|}{} & \multicolumn{1}{|c|}{} &\multicolumn{1}{c|}{}&
    testA     & testB    \\
    
    \midrule
    \midrule
    
    \multicolumn{1}{c}{\multirow{6}{*}{REC}} &
    \multicolumn{1}{|c|}{\multirow{3}{*}{r50}} &

     RGB & 
    \textbf{86.92} & 54.12    \\
    \multicolumn{1}{c|}{} & \multicolumn{1}{|c|}{} &
     Depth &
    52.04 & 22.03      \\
    \multicolumn{1}{c|}{} & \multicolumn{1}{|c|}{} &
     RGB-D &
    84.22 & 48.12  \\
    
    \cline{2-5}
    \multicolumn{1}{c|}{} &
    \multicolumn{1}{|c|}{\multirow{3}{*}{r101}\rule{0pt}{2.5ex}} & 
     RGB & 
    85.62 & \textbf{55.97}      \\
    \multicolumn{1}{c|}{} & \multicolumn{1}{|c|}{} &
     Depth &
    44.03 & 17.46   \\
    \multicolumn{1}{c|}{} & \multicolumn{1}{|c|}{} &
     RGB-D &
   81.19 & 45.68 \\
    \midrule
    \midrule
    
    \multicolumn{1}{c}{\multirow{6}{*}{RES}} &
    \multicolumn{1}{|c|}{\multirow{3}{*}{r50}} &
    
     RGB & 
    \textbf{85.46} & \textbf{61.49}   \\
    \multicolumn{1}{c|}{} & \multicolumn{1}{|c|}{} &
     Depth &
    50.62 & 25.19      \\
    \multicolumn{1}{c|}{} & \multicolumn{1}{|c|}{} &
     RGB-D &
    81.16 & 52.98  \\
    
    \cline{2-5}
    \multicolumn{1}{c|}{} &
    \multicolumn{1}{|c|}{\multirow{3}{*}{r101}\rule{0pt}{2.5ex}} & 
     RGB & 
    83.89 & 60.72   \\
    \multicolumn{1}{c|}{} & \multicolumn{1}{|c|}{} &
     Depth &
    48.86 & 24.36     \\
    \multicolumn{1}{c|}{} & \multicolumn{1}{|c|}{} &
     RGB-D &
    78.07& 49.75 \\
    
    \bottomrule
  \end{tabular}
  \end{minipage}
\end{table}

\paragraph{Training Settings.}
\label{settings}
We train the RefTR model \cite{li2021referring} on the RoboRefIt by using 4 NVIDIA Geforce RTX 3090 GPU with synchronous training. The initial learning rate of the model is set to $1e{-5}$, the batch size is set to 16 and training is terminated at the 90th epoch. The ResNet-50 or ResNet-101 \cite{he2016deep} are used as image backbone without pre-train. In the process of image preprocessing, the scale of the longest edge of the image is 640 pixel and the shape of the mask array is \emph{480$\times$640$\times$1}. According to the statistic of the expressions in RoboRefIt, we set the maximum length of context sentence to 30. And each sentence have only one query phrase that is itself. Especially, in addition to inputting only RGB image as input of the model, we also feed the depth image into the network for ablation experiments. The depth map can directly combine with the RGB image for learning tasks \cite{xie2020best} or be transformed into point cloud for segmentation \cite{papon2013voxel, cui2021deep}. In this work, we use the most naive way to directly concatenate the depth image with the RGB image, hence the dimension of the RGB-D input is \emph{480$\times$640$\times$4}. Also, in the contrast, we input a single depth image with a dimension of \emph{480$\times$640$\times$1} to the network to predict results. 

\paragraph{Results on RoboRefIt.} Following prior work \cite{li2021referring}, the evaluation metric for the REC task is precision$@0.5$ (prec$@0.5$), which measures the percentage of test images with an intersection-over-union (IoU) score higher than the threshold $0.5$. The IoU score calculates intersection regions over union regions of the predicted bounding box and the ground truth. For the RES task, the mean IoU (mIoU) between the predicted segmentation mask and the ground truth is used as the evaluation metric. As is shown in Table \ref{two:experiment}, we adopt the RefTR model \cite{li2021referring} to benchmark the RoboRefIt dataset and conduct REC and RES experiments in testA and testB set. We conduct ablation experiments with different image backbone or different image inputs. According to the results, we use the ResNet50 as the image backbone and the RGB input in the VL-Grasp stage \uppercase\expandafter{\romannumeral1}.

\subsection{Real Robot Experiments}
Real robotic grasping experiments are conducted to demonstrate the performance and practicability of the VL-Grasp. The experimental equipments include a UR3 robotic arm, an OnRobot RG2 gripper, an Intel Realsense D455 camera and a Geforce GTX 1050ti GPU. To move to different scenes for grasping, we install the robotic arm on a Oasis 300C mobile robot. The VL-Grasp trains the visual grounding network with ResNet50 backbone and RGB input at RoboRefIt, and trains the GPD network at GraspNet-1Billion \cite{fang2020graspnet}. For more training setting about FGC-GraspNet, please refer to \cite{lu2022hybrid}. In real robot experiments, we set up practical object scenes and natural language commands that are similar to the datasets. In the following, three experiments are designed to evaluate the performance of the VL-Grasp.

\begin{table}[t]
\begin{minipage}{0.5\textwidth}
\centering
\caption{Comparative Results on Object Categories.}
\label{table:object}
\begin{tabular}{c|c|c|c}
\toprule

\multicolumn{1}{c|}{Object}& 
\multicolumn{1}{|c|}{Public}&
\multicolumn{1}{|c|}{GPD-unseen}&
\multicolumn{1}{|c}{Success rate} \\
\hline

\multicolumn{1}{c|}{Banana}& 
\multicolumn{1}{c|}{\checkmark}&
\multicolumn{1}{c|}{}&
\multicolumn{1}{c}{7/10} \\

\multicolumn{1}{c|}{Apple}& 
\multicolumn{1}{c|}{\checkmark}&
\multicolumn{1}{c|}{}&
\multicolumn{1}{c}{10/10} \\

\multicolumn{1}{c|}{Elephant Model}& 
\multicolumn{1}{c|}{\checkmark}&
\multicolumn{1}{c|}{}&
\multicolumn{1}{c}{8/10} \\

\multicolumn{1}{c|}{Toothpaste Bottle}& 
\multicolumn{1}{c|}{\checkmark}&
\multicolumn{1}{c|}{}&
\multicolumn{1}{c}{9/10} \\

\multicolumn{1}{c|}{Dinosaur Model}& 
\multicolumn{1}{c|}{}&
\multicolumn{1}{c|}{\checkmark}&
\multicolumn{1}{c}{6/10} \\

\multicolumn{1}{c|}{Tennis}& 
\multicolumn{1}{c|}{}&
\multicolumn{1}{c|}{\checkmark}&
\multicolumn{1}{c}{7/10} \\

\multicolumn{1}{c|}{Vitamin Bottle}& 
\multicolumn{1}{c|}{}&
\multicolumn{1}{c|}{\checkmark}&
\multicolumn{1}{c}{9/10} \\

\multicolumn{1}{c|}{Cone}& 
\multicolumn{1}{c|}{}&
\multicolumn{1}{c|}{\checkmark}&
\multicolumn{1}{c}{7/10} \\

\bottomrule

\end{tabular}
\end{minipage}
\end{table}

\subsubsection{Comparative Results on Object Categories}
In this section, we demonstrate the generalization ability of the VL-Grasp in real situation and compare the model performance on different objects.
The experimental objects are divided into two categories. The first category is seen in both of RoboRefIt and GraspNet-1Billion \cite{fang2020graspnet}, named public objects. And the second category is seen in RoboRefIt but unseen in GraspNet-1Billion \cite{fang2020graspnet}, named GPD-unseen objects. We choose several objects with representative shape and set up the physical environment randomly. Each object with table scenes is tested with 10 attempts. According to the results in Table \ref{table:object}, the VL-Grasp can be generalized to grasp diverse object categories that belong to RoboRefIt. Although the object is unseen for the trained GPD model, the VL-Grasp can also handle with similar-shape objects.

\begin{table}[t]
\begin{minipage}{0.5\textwidth}
\centering
\caption{Comparative Results on Different Scenes.}
\label{table:scene}
\begin{tabular}{c|c|c|c|c}
\toprule

\multicolumn{1}{c|}{Object}& 
\multicolumn{1}{|c|}{Table}&
\multicolumn{1}{|c|}{Sofa}&
\multicolumn{1}{|c|}{Shelf} &
\multicolumn{1}{c}{Success rate} \\
\hline

\multicolumn{1}{c|}{\multirow{3}{*}{Toothpaste Bottle}}& 
\multicolumn{1}{c|}{\checkmark}&
\multicolumn{1}{c|}{}&
\multicolumn{1}{c|}{}&
\multicolumn{1}{c}{9/10} \\

\multicolumn{1}{c|}{}& 
\multicolumn{1}{c|}{}&
\multicolumn{1}{c|}{\checkmark}&
\multicolumn{1}{c|}{}&
\multicolumn{1}{c}{7/10} \\

\multicolumn{1}{c|}{}& 
\multicolumn{1}{c|}{}&
\multicolumn{1}{c|}{}&
\multicolumn{1}{c|}{\checkmark}&
\multicolumn{1}{c}{8/10}\\

\hline

\multicolumn{1}{c|}{\multirow{3}{*}{Vitamin Bottle}}& 
\multicolumn{1}{c|}{\checkmark}&
\multicolumn{1}{c|}{}&
\multicolumn{1}{c|}{}&
\multicolumn{1}{c}{9/10}\\

\multicolumn{1}{c|}{}& 
\multicolumn{1}{c|}{}&
\multicolumn{1}{c|}{\checkmark}&
\multicolumn{1}{c|}{}&
\multicolumn{1}{c}{7/10} \\

\multicolumn{1}{c|}{}& 
\multicolumn{1}{c|}{}&
\multicolumn{1}{c|}{}&
\multicolumn{1}{c|}{\checkmark}&
\multicolumn{1}{c}{8/10} \\
\hline

\multicolumn{1}{c|}{\multirow{3}{*}{Cone}}& 
\multicolumn{1}{c|}{\checkmark}&
\multicolumn{1}{c|}{}&
\multicolumn{1}{c|}{}&
\multicolumn{1}{c}{7/10}\\

\multicolumn{1}{c|}{}& 
\multicolumn{1}{c|}{}&
\multicolumn{1}{c|}{\checkmark}&
\multicolumn{1}{c|}{}&
\multicolumn{1}{c}{5/10} \\

\multicolumn{1}{c|}{}& 
\multicolumn{1}{c|}{}&
\multicolumn{1}{c|}{}&
\multicolumn{1}{c|}{\checkmark}&
\multicolumn{1}{c}{7/10} \\
\hline

\multicolumn{1}{c|}{\multirow{3}{*}{Shampoo Bottle}}& 
\multicolumn{1}{c|}{\checkmark}&
\multicolumn{1}{c|}{}&
\multicolumn{1}{c|}{}&
\multicolumn{1}{c}{6/10}\\

\multicolumn{1}{c|}{}& 
\multicolumn{1}{c|}{}&
\multicolumn{1}{c|}{\checkmark}&
\multicolumn{1}{c|}{}&
\multicolumn{1}{c}{7/10} \\

\multicolumn{1}{c|}{}& 
\multicolumn{1}{c|}{}&
\multicolumn{1}{c|}{}&
\multicolumn{1}{c|}{\checkmark}&
\multicolumn{1}{c}{7/10} \\
\hline

\multicolumn{1}{c|}{Overall}& 
\multicolumn{1}{c|}{\checkmark}&
\multicolumn{1}{c|}{\checkmark}&
\multicolumn{1}{c|}{\checkmark}&
\multicolumn{1}{c}{72.5\%} \\

\bottomrule

\end{tabular}
\end{minipage}
\end{table}

\subsubsection{Comparative Results on Different Scenes}
Previous works \cite{chen2021joint, ding2022visual, shridhar2020ingress, zhang2021invigorate} solve the problem that grasping objects from a restrictive single perspective. In this section, we set up different indoor scenes at different observation views for interactive grasping. The previous approaches are no longer applicable to such interaction scenarios. The experimental results in Table \ref{table:scene} validate that the VL-Grasp still has high adaptability and efficient grasp success rate in variant environments. The average success rate in different scenes reach 72.5\%.

\begin{table}[t]
\begin{minipage}{0.5\textwidth}
\centering
\caption{Ablation Results on Point Cloud Filter.}
\label{table:ablation}
\begin{tabular}{c|c|c|c}
\toprule

\multicolumn{1}{c|}{Object}& 
\begin{tabular}[c]{@{}c@{}}Pre filter \\ with mask\end{tabular}&
\multicolumn{1}{|c|}{Point cloud filter}&
\multicolumn{1}{|c}{Success rate} \\

\hline

\multicolumn{1}{c|}{\multirow{3}{*}{Lion}}& 
\multicolumn{1}{c|}{-}&
\multicolumn{1}{c|}{-}&
\multicolumn{1}{c}{2/10} \\

\multicolumn{1}{c|}{}& 
\multicolumn{1}{c|}{\checkmark}&
\multicolumn{1}{c|}{}&
\multicolumn{1}{c}{0/10} \\

\multicolumn{1}{c|}{}& 
\multicolumn{1}{c|}{}&
\multicolumn{1}{c|}{\checkmark}&
\multicolumn{1}{c}{7/10} \\

\hline

\multicolumn{1}{c|}{\multirow{3}{*}{Orange}}& 
\multicolumn{1}{c|}{-}&
\multicolumn{1}{c|}{-}&
\multicolumn{1}{c}{2/10} \\

\multicolumn{1}{c|}{}& 
\multicolumn{1}{c|}{\checkmark}&
\multicolumn{1}{c|}{}&
\multicolumn{1}{c}{1/10} \\

\multicolumn{1}{c|}{}& 
\multicolumn{1}{c|}{}&
\multicolumn{1}{c|}{\checkmark}&
\multicolumn{1}{c}{8/10} \\

\bottomrule

\end{tabular}
\end{minipage}
\end{table}

\begin{figure*}[t]
  \centering
  \includegraphics[width=\textwidth]{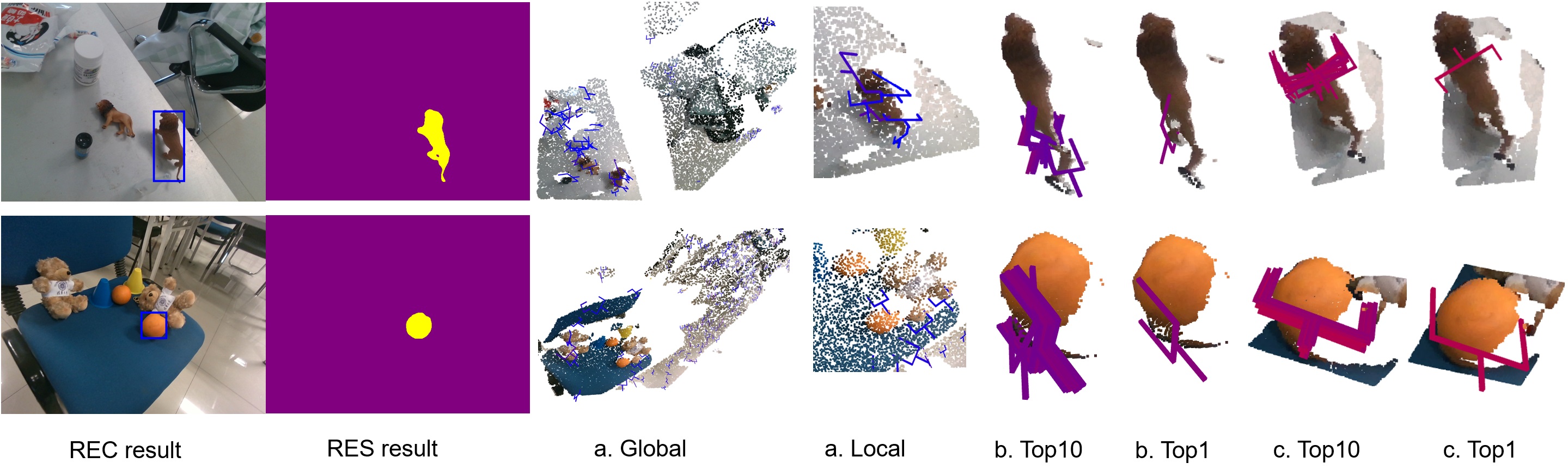}
  \caption{Visualization of Ablation Results. The a refers to without pre-filter. The b refers to pre-filter with mask. The c refers to our point cloud filter.}
  \label{fig:ablation}
\end{figure*}

\subsubsection{Ablation Results on Point Cloud Filter}
The point cloud filter module is designed to generate object-level point cloud. We conduct ablation experiments to demonstrate the effectiveness of the point cloud filter. There are two control experiments: without pre-filter and pre-filter with mask. As is shown in \figref{fig:ablation}, the performance of predicting grasp poses under the point cloud filter is far superior to the other two filtering strategies. Without pre-filter module, the GPD model is prone to predict heavy-duty inaccurate grasp poses in cluttered background points and can't focus on the reasoning of the target region. Pre-filter with mask only leaves the point cloud information of the target, which makes the GPD model unable to judge the position and posture of the object with its environment surroundings. The results in Table \ref{table:ablation} illustrate that the point cloud filter module enhance the success rate of grasping in real environment.

\section{Conclusion}

This paper proposes a robotic interactive grasp policy based on multi-modal perception and reasoning, VL-Grasp. In the meanwhile, a new visual grounding dataset, RoboRefIt, is built to fills the vacancy of visual grounding researches in the robot field. The VL-Grasp combines the visual grounding and 6-Dof grasp pose detection network, and trains the visual grounding network in RoboRefIt. Consequently, the VL-Grasp has expansibility and adaptability in practical grasping. In the future, we will explore how to achieve the interactive grasp in open world with unseen objects.

\section{Acknowledgment}

This work was supported by the research fund under Grant No.2019GQG0001 from the Institute for Guo Qiang,Tsinghua University.

{
\small

\bibliographystyle{plain}
\bibliography{root}

}

\addtolength{\textheight}{-12cm}   


\end{document}